\documentclass{article}

\usepackage[final,nonatbib]{nips_2017}

\usepackage[utf8]{inputenc} % allow utf-8 input
\usepackage[T1]{fontenc}    % use 8-bit T1 fonts
\usepackage{hyperref}       % hyperlinks
\usepackage{url}            % simple URL typesetting
\usepackage{booktabs}       % professional-quality tables
\usepackage{amsfonts}       % blackboard math symbols
\usepackage{nicefrac}       % compact symbols for 1/2, etc.
\usepackage{microtype}      % microtypography
\usepackage{comment}

\usepackage{times}
\usepackage{graphicx} % more modern
\usepackage{subfigure}
\usepackage{wrapfig}
\usepackage{lipsum}
\usepackage{adjustbox}

\usepackage{epstopdf}
\usepackage{amsmath}
\usepackage{amssymb}
\usepackage{amsthm}
\newtheorem{theorem}{Theorem}[section]

\newtheorem{lemma}[theorem]{Lemma}
\usepackage{color}
\usepackage{algorithm}
\usepackage{algorithmic}

\usepackage{hyperref}
\usepackage{mathtools}
\usepackage{bbm}
\def\etal{{et al.\/}\,}
\usepackage{caption}
\newcommand\ExerciseCaption[1]{\captionsetup{font=footnotesize}\caption{#1}}

\title{Learning to Prune Deep Neural Networks via \\ Layer-wise Optimal Brain Surgeon}

\author{
  Xin Dong \\
  Nanyang Technological University, Singapore \\
  \texttt{n1503521a@e.ntu.edu.sg} \\
   \And
   Shangyu Chen \\
   Nanyang Technological University, Singapore \\
   \texttt{schen025@e.ntu.edu.sg} \\
   \AND
   Sinno Jialin Pan \\
   Nanyang Technological University, Singapore \\
   \texttt{sinnopan@ntu.edu.sg} \\
}

\begin{document}

\maketitle

\begin{abstract}
How to develop slim and accurate deep neural networks has become crucial for real-world applications, especially for those employed in embedded systems. Though previous work along this research line has shown some promising results, most existing methods either fail to significantly compress a well-trained deep network or require a heavy retraining process for the pruned deep network to re-boost its prediction performance. In this paper, we propose a new layer-wise pruning method for deep neural networks. In our proposed method, parameters of each individual layer are pruned independently based on second order derivatives of a layer-wise error function with respect to the corresponding parameters. We prove that the final prediction performance drop after pruning is bounded by a linear combination of the reconstructed errors caused at each layer. %Therefore, there is a guarantee that one only needs to perform a light retraining process on the pruned network to resume its original prediction performance.
By controlling layer-wise errors properly, one only needs to perform a light retraining process on the pruned network to resume its original prediction performance.
We conduct extensive experiments on benchmark datasets to demonstrate the effectiveness of our pruning method compared with several state-of-the-art baseline methods. Codes of our work are released at: \texttt{https://github.com/csyhhu/L-OBS}.
\end{abstract}

\section{Introduction}
\label{introduction}
%Inspired by the structure of human brain, deep neural networks have become a ubiquitous technology in many real-world applications~\cite{deepnature}.
Intuitively, deep neural networks~\cite{deepnature} can approximate predictive functions of arbitrary complexity well when they are of a huge amount of parameters, i.e., a lot of layers and neurons. In practice, the size of deep neural networks has been being tremendously increased, from LeNet-5 with less than 1M parameters~\cite{lenet5} to VGG-16 with 133M parameters~\cite{vgg16}. Such a large number of parameters not only make deep models memory intensive and computationally expensive, but also urge researchers to dig into redundancy of deep neural networks. On one hand, in neuroscience, recent studies point out that there are significant redundant neurons in human brain, and memory may have relation with vanishment of specific synapses~\cite{de2017ultrastructural}. On the other hand, in machine learning, both theoretical analysis and empirical experiments have shown the evidence of redundancy in several deep models~\cite{r1,nettrim}. Therefore, it is possible to compress deep neural networks without or with little loss in prediction by pruning parameters with carefully designed criteria.

However, finding an optimal pruning solution is NP-hard because the search space for pruning is exponential in terms of parameter size. Recent work mainly focuses on developing efficient algorithms to obtain a near-optimal pruning solution~\cite{reed1993pruning,gong2014compressing,hannips,sun2016sparsifying,dynamic}. A common idea behind most exiting approaches is to select parameters for pruning based on certain criteria, such as increase in training error, magnitude of the parameter values, etc. As most of the existing pruning criteria are designed heuristically, there is no guarantee that prediction performance of a deep neural network can be preserved after pruning. Therefore, a time-consuming retraining process is usually needed to boost the performance of the trimmed neural network.

Instead of consuming efforts on a whole deep network, a layer-wise pruning method, Net-Trim, was proposed to learn sparse parameters by minimizing reconstructed error for each individual layer~\cite{nettrim}. A theoretical analysis is provided that the overall performance drop of the deep network is bounded by the sum of reconstructed errors for each layer. In this way, the pruned deep network has a theoretical guarantee on its %performance
error. However, as Net-Trim adopts $\ell_1$-norm to induce sparsity for pruning, it fails to obtain high compression ratio compared with other methods~\cite{hannips,dynamic}.

In this paper, we propose a new layer-wise pruning method for deep neural networks, aiming to achieve the following three goals: 1) For each layer, parameters can be highly compressed after pruning, while the reconstructed error is small. 2) There is a theoretical guarantee on the overall prediction performance of the pruned deep neural network in terms of reconstructed errors for each layer. 3) After the deep network is pruned, only a light retraining process is required to resume its original prediction performance.

To achieve our first goal, we borrow an idea from some classic pruning approaches for shallow neural networks, such as optimal brain damage (OBD)~\cite{obd} and optimal brain surgeon (OBS)~\cite{obs}. These classic methods approximate a change in the error function via functional Taylor Series, and identify unimportant weights based on second order derivatives. Though these approaches have proven to be effective for shallow neural networks, it remains challenging to extend them for deep neural networks because of the high computational cost on computing second order derivatives, i.e., the inverse of the Hessian matrix over all the parameters. In this work, as we restrict the computation on second order derivatives w.r.t. the parameters of each individual layer only, i.e., the Hessian matrix is only over parameters for a specific layer, the computation becomes tractable. Moreover, we utilize characteristics of back-propagation for fully-connected layers in well-trained deep networks to further reduce computational complexity of the inverse operation of the Hessian matrix.

To achieve our second goal, based on the theoretical results in~\cite{nettrim}, we provide a proof on the bound of performance drop before and after pruning in terms of the reconstructed errors for each layer. With such a layer-wise pruning framework using second-order derivatives for trimming parameters for each layer, we empirically show that after significantly pruning parameters, there is only a little drop of prediction performance compared with that before pruning. Therefore, only a light retraining process is needed to resume the performance, which achieves our third goal.

The contributions of this paper are summarized as follows. 1) We propose a new layer-wise pruning method for deep neural networks, which is able to significantly trim networks and preserve the prediction performance of networks after pruning with a theoretical guarantee. In addition, with the proposed method, a time-consuming retraining process for re-boosting the performance of the pruned network is waived. 2) We conduct extensive experiments to verify the effectiveness of our proposed method compared with several state-of-the-art approaches.

\section{Related Works and Preliminary}
\label{relatedworks}
Pruning methods have been widely used for model compression in early neural networks~\cite{reed1993pruning} and modern deep neural networks~\cite{nettrim,gong2014compressing,hannips,sun2016sparsifying,dynamic}. In the past, with relatively small size of training data, pruning is crucial to avoid overfitting. Classical methods include OBD and OBS. These methods aim to prune parameters with the least increase of error approximated by second order derivatives. However, computation of the Hessian inverse over all the parameters is expensive. In OBD, the Hessian matrix is restricted to be a diagonal matrix to make it computationally tractable. However, this approach implicitly assumes parameters have no interactions, which may hurt the pruning performance. %Different from OBD, OBS makes use of the full Hessian matrix for pruning, which obtains better performance while is much more computational expensive.
Different from OBD, OBS makes use of the full Hessian matrix for pruning. It obtains better performance while is much more computationally expensive even using Woodbury matrix identity~\cite{kailath1980linear}, which is an iterative method to compute the Hessian inverse. For example, using OBS on VGG-16 naturally requires to compute inverse of the Hessian matrix with a size of $133\mbox{M}\times 133\mbox{M}$.

Regarding pruning for modern deep models, Han~\etal~\cite{hannips} proposed to delete unimportant parameters based on magnitude of their absolute values, and retrain the remaining ones to recover the original prediction performance. This method achieves considerable compression ratio in practice. However, as pointed out by pioneer research work~\cite{obd,obs}, parameters with low magnitude of their absolute values can be necessary for low error. Therefore, magnitude-based approaches may eliminate wrong parameters, resulting in a big prediction performance drop right after pruning, and poor robustness before retraining~\cite{r2}. Though some variants have tried to find better magnitude-based criteria~\cite{cuhk,li2016pruning}, the significant drop of prediction performance after pruning still remains. To avoid pruning wrong parameters, Guo~\etal~\cite{dynamic} introduced a mask matrix to indicate the state of network connection for dynamically pruning after each gradient decent step. Jin~\etal~\cite{ith} proposed an iterative hard thresholding approach to re-activate the pruned parameters after each pruning phase.

Besides Net-trim, which is a layer-wise pruning method discussed in the previous section, there is some other work proposed to induce sparsity or low-rank approximation on certain layers for pruning~\cite{lowrank,lasso}. However, as the $\ell_0$-norm or the $\ell_1$-norm sparsity-induced regularization term increases difficulty in optimization, the pruned deep neural networks using these methods either obtain much smaller compression ratio~\cite{nettrim} compared with direct pruning methods or require retraining of the whole network to prevent accumulation of errors~\cite{sun2016sparsifying}.

\noindent{\bf Optimal Brain Surgeon} As our proposed layer-wise pruning method is an extension of OBS on deep neural networks, we briefly review the basic of OBS here. Consider a network in terms of parameters $\mathbf{w}$ trained to a local minimum in error. The functional Taylor series of the error w.r.t. $\mathbf{w}$ is:
%\begin{equation}\label{eq:obs:taylor}
$\delta E = \left(\frac{\partial E}{\partial \mathbf{w}}\right)^\top\delta \mathbf{w} + \frac{1}{2}\delta {\mathbf{w}}^\top \mathbf{H} \delta \mathbf{w} + O\left(\|\delta \mathbf{w}\|^3\right)$,
%\end{equation}
where $\delta$ denotes a perturbation of a corresponding variable, $\mathbf{H} \equiv \partial^2E/\partial \mathbf{w}^2\in\mathbb{R}^{m\times m}$ is the Hessian matrix, where $m$ is the number of parameters, and $O(\|\delta \boldsymbol{\Theta}_l\|^3)$ is the third and all higher order terms. For a network trained to a local minimum in error, the first term vanishes, and the term $O(\|\delta \boldsymbol{\Theta}_l\|^3)$ can be ignored. In OBS, the goal is to set one of the parameters to zero, denoted by $w_q$ (scalar), to minimize $\delta E$ in each pruning iteration. The resultant optimization problem is written as follows,
\begin{equation}
\label{obs:formulaton}
	\min_q \frac{1}{2}\delta {\mathbf{w}}^\top \mathbf{H}\delta \mathbf{w}, \;\; \mbox{s.t. }\; \mathbf{e}_q^\top \delta \mathbf{w} + \mathbf{w}_q = 0,
\end{equation}
where $\mathbf{e}_q$ is the unit selecting vector whose $q$-th element is 1 and otherwise 0. As shown in~\cite{lag}, the optimization problem \eqref{obs:formulaton} can be solved by the Lagrange multipliers method. Note that a computation bottleneck of OBS is to calculate and store the non-diagonal Hesssian matrix and its inverse, which makes it impractical on pruning deep models which are usually of a huge number of parameters.

\section{Layer-wise Optimal Brain Surgeon}
\subsection{Problem Statement}
%In the following, we present how to extend OBS, one of the most popular second-derivative methods for sallow neural networks, to prune deep neural networks. A whole deep neural network can be considered as a highly nonlinear and non-convex mapping function from input features to output labels. Computing the Hessian matrix over all the parameters in such deep networks is computationally intractable. Therefore, motivated by Net-trim~\cite{nettrim}, we propose to decompose the pruning problem over the whole deep neural network into a set of layer-wise pruning problems independently with theoretical analysis on accumulated errors across layers.

Given a training set of $n$ instances, $\{(\mathbf{x}_j,y_j)\}_{j=1}^n$, and a well-trained deep neural network of $L$ layers (excluding the input layer)\footnote{For simplicity in presentation, we suppose the neural network is a feed-forward (fully-connected) network. In Section~\ref{parallel}, we will show how to extend our method to filter layers in Convolutional Neural Networks.}. Denote the input and the output of the whole deep neural network by $\mathbf{X}\!=\![\mathbf{x}_1,...,\mathbf{x}_n]\!\in\!\mathbb{R}^{d\times n}$ and $\mathbf{Y}\!\in\!\mathbb{R}^{n\times 1}$, respectively. For a layer $l$, we denote the input and output of the layer by $\mathbf{Y}^{l-1}\!=\![\mathbf{y}_1^{l-1},...,\mathbf{y}_n^{l-1}]\!\in\!\mathbb{R}^{m_{l-1}\times n}$ and $\mathbf{Y}^{l}\!=\![\mathbf{y}_1^{l},...,\mathbf{y}_n^{l}]\!\in\!\mathbb{R}^{m_{l}\times n}$, respectively, where $\mathbf{y}^l_i$ can be considered as a representation of $\mathbf{x}_i$ in layer $l$, and $\mathbf{Y}^0=\mathbf{X}$, $\mathbf{Y}^L=\mathbf{Y}$, and $m_0=d$. Using one forward-pass step, we have $\mathbf{Y}^l\!=\!\sigma(\mathbf{Z}^{l})$, where $\mathbf{Z}^{l}\!=\!{\mathbf{W}_l}^\top \mathbf{Y}^{l-1}$ with $\mathbf{W}_l\!\in\! \mathbb{R}^{m_{l-1}\times m_{l}}$ being the matrix of parameters for layer $l$, and $\sigma(\cdot)$ is the activation function. For convenience in presentation and proof, we define the activation function $\sigma(\cdot)$ as the rectified linear unit (ReLU)~\cite{relu}. We further denote by $\boldsymbol{\Theta}_l\!\in\! \mathbb{R}^{m_{l-1}m_{l}\times 1}$ the vectorization of $\mathbf{W}_l$. For a well-trained neural network, $\mathbf{Y}^l$, $\mathbf{Z}^l$ and $\boldsymbol{\Theta}_l^*$ are all fixed matrixes and contain most information of the neural network. The goal of pruning is to set the values of some elements in $\boldsymbol{\Theta}_l$ to be zero.

\subsection{Layer-Wise Error}
\label{layerwise}
During layer-wise pruning in layer $l$, the input $\mathbf{Y}^{l-1}$ is fixed as the same as the well-trained network. Suppose we set the $q$-th element of $\boldsymbol{\Theta}_l$, denoted by $\boldsymbol{\Theta}_{l_{[q]}}$, to be zero, and get a new parameter vector, denoted by $\boldsymbol{\hat{\Theta}}_{l}$. With $\mathbf{Y}^{l-1}$, we obtain a new output for layer $l$, denoted by $\mathbf{\hat{Y}}^{l}$. Consider the root of mean square error between $\mathbf{\hat{Y}}^{l}$ and $\mathbf{Y}^{l}$ over the whole training data as the layer-wise error:
\begin{equation}
\label{eq:epsilon}
\varepsilon^l = \sqrt{\frac{1}{n}\sum_{j=1}^n\left((\mathbf{\hat{y}}_{j}^{l}-\mathbf{y}_{j}^{l})^\top (\mathbf{\hat{y}}_{j}^{l}-\mathbf{y}_{j}^{l})\right)}
              =  \frac{1}{\sqrt{n}}\|\mathbf{\hat{Y}}^{l}-\mathbf{Y}^{l}\|_F,
\end{equation}
where $\|\cdot\|_F$ is the Frobenius Norm. Note that for any single parameter pruning, one can compute its error $\varepsilon^l_q$, where $1\leq q \leq m_{l-1}m_l$, and use it as a pruning criterion. This idea has been adopted by some existing methods~\cite{r2}. However, in this way, for each parameter at each layer, one has to pass the whole training data once to compute its error measure, which is very computationally expensive. A more efficient approach is to make use of the second order derivatives of the error function to help identify importance of each parameter.

We first define an error function $E(\cdot)$ as %with respect to $\mathbf{Z}^l$ (fixed outcome of weighted sum before performing the activation function $ReLU$ from the well-trained neural network) as
\begin{equation}\label{eq:errfun}
  E^l = E(\mathbf{\hat{Z}}^l) = \frac{1}{n}\left\|\mathbf{\hat{Z}}^l - \mathbf{Z}^{l}\right\|^2_F,
\end{equation}
where $\mathbf{Z}^l$ is outcome of the weighted sum operation right before performing the activation function $\sigma(\cdot)$ at layer $l$ of the well-trained neural network, and $\mathbf{\hat{Z}}^l$ is outcome of the weighted sum operation after pruning at layer $l$ . Note that $\mathbf{Z}^l$ is considered as the desired output of layer $l$ before activation.
The following lemma shows that the layer-wise error is bounded by the error defined in \eqref{eq:errfun}.
\begin{lemma}
\label{th11}
	With the error function \eqref{eq:errfun} and $\mathbf{Y}^{l} = \sigma(\mathbf{Z}^{l})$, the following holds: $\varepsilon^l \leq \sqrt{E(\mathbf{\hat{Z}}^l)}$.
\end{lemma}
Therefore, to find parameters whose deletion (set to be zero) minimizes \eqref{eq:epsilon} can be translated to find parameters those deletion minimizes the error function \eqref{eq:errfun}. Following~\cite{obd,obs}, the error function can be approximated by functional Taylor series as follows,
\begin{equation}\label{eq:taylor}
	 E(\mathbf{\hat{Z}}^l) -  E(\mathbf{Z}^l) = \delta E^l = \left(\frac{\partial E^l}{\partial \boldsymbol{\Theta}_l}\right)^\top\delta \boldsymbol{\Theta}_l + \frac{1}{2}\delta {\boldsymbol{\Theta}_l}^\top \mathbf{H}_l \delta \boldsymbol{\Theta}_l + O\left(\|\delta \boldsymbol{\Theta}_l\|^3\right),
\end{equation}
where $\delta$ denotes a perturbation of a corresponding variable, $\mathbf{H}_l \equiv \partial^2E^l/\partial {\boldsymbol{\Theta}_l}^2$ is the Hessian matrix w.r.t. $\boldsymbol{\Theta}_l$, and $O(\|\delta \boldsymbol{\Theta}_l\|^3)$ is the third and all higher order terms. It can be proven that with the error function defined in \eqref{eq:errfun}, the first (linear) term  $\left.\frac{\partial E^l}{\partial \boldsymbol{\Theta}_l}\!\right|\!_{\boldsymbol{\Theta}_l=\boldsymbol{\Theta}_l^*}$ and $O(\|\delta \boldsymbol{\Theta}_l\|^3)$ are equal to $0$.

Suppose every time one aims to find a parameter $\boldsymbol{\Theta}_{l_{[q]}}$ to set to be zero such that the change $\delta E^l$ is minimal. Similar to OBS, we can formulate it as the following optimization problem:
\begin{equation}
\label{formulaton}
	\min_q \frac{1}{2}\delta {\boldsymbol{\Theta}_l}^\top \mathbf{H}_l\delta \boldsymbol{\Theta}_l, \;\; \mbox{s.t. }\; \mathbf{e}_q^\top \delta \boldsymbol{\Theta}_l + \boldsymbol{\Theta}_{l_{[q]}} = 0,
\end{equation}
where $\mathbf{e}_q$ is the unit selecting vector whose $q$-th element is 1 and otherwise 0. By using the Lagrange multipliers method as suggested in~\cite{lag}, we obtain the closed-form solutions of the optimal parameter pruning and the resultant minimal change in the error function as follows,
\begin{eqnarray}
	\delta \boldsymbol{\Theta}_l = -\frac{\boldsymbol{\Theta}_{l_{[q]}}}{[\mathbf{H}_l^{-1}]_{qq}}\mathbf{H}_l^{-1}\mathbf{e}_q, \;\mbox{ and }\;
	L_q = \delta E^l = \frac{1}{2}\frac{({\boldsymbol{\Theta}_{l_{[q]}}})^2}{[\mathbf{H}_l^{-1}]_{qq}}. \label{changel}
\end{eqnarray}
Here $L_q$ is referred to as the sensitivity of parameter $\boldsymbol{\Theta}_{l_{[q]}}$. Then we select parameters to prune based on their sensitivity scores instead of their magnitudes. As mentioned in section~\ref{relatedworks}, magnitude-based criteria which merely consider the numerator in \eqref{changel} is a poor estimation of sensitivity of parameters. Moreover, in \eqref{changel}, as the inverse Hessian matrix over the training data is involved, it is able to capture data distribution when measuring sensitivities of parameters.

After pruning the parameter, $\boldsymbol{\Theta}_{l_{[q]}}$, with the smallest sensitivity, the parameter vector is updated via $\boldsymbol{\hat{\Theta}}_l \!=\! \boldsymbol{\Theta}_l \!+\! \delta \boldsymbol{\Theta}_l$. With Lemma~\ref{th11} and \eqref{changel}, we have that the layer-wise error for layer $l$ is bounded by
\begin{eqnarray}\label{onelayer}
	\varepsilon^l_{q} \leq \sqrt{E(\mathbf{\hat{Z}}^l)} = \sqrt{E(\mathbf{\hat{Z}}^l) - E(\mathbf{Z}^{l})} = \sqrt{\delta E^l} = \frac{\vert\boldsymbol{\Theta}_{l_{[q]}}\vert}{\sqrt{2[\mathbf{H}_l^{-1}]_{qq}}}.
\end{eqnarray}
Note that first equality is obtained because of the fact that $E(\mathbf{Z}^{l})=0$. It is worth to mention that though we merely focus on layer $l$, the Hessian matrix is still a square matrix with size of $m_{l-1}m_l\times m_{l-1}m_l$. However, we will show how to significantly reduce the computation of $\mathbf{H}^{-1}_l$ for each layer in Section~\ref{parallel}.

\subsection{Layer-Wise Error Propagation and Accumulation}
So far, we have shown how to prune parameters for each layer, and estimate their introduced errors independently. However, our aim is to control the consistence of the network's final output $\mathbf{Y}^L$ before and after pruning. To do this, in the following, we show how the layer-wise errors propagate to final output layer, and the accumulated error over multiple layers will \textbf{not} explode.
\begin{theorem}
\label{th1}
Given a pruned deep network via layer-wise pruning introduced in Section~\ref{layerwise}, each layer has its own layer-wise error $\varepsilon^l$ for $1\leq l\leq L$, then the accumulated error of ultimate network output $\tilde{\varepsilon}^L = \frac{1}{\sqrt{n}}\|\mathbf{\tilde{Y}}^{L}-\mathbf{Y}^{L}\|_F$ obeys:
\begin{equation}
\label{finallayer}
	\tilde{\varepsilon}^L\leq \sum_{k=1}^{L-1}\left(\prod_{l=k+1}^L\|\boldsymbol{\hat{\Theta}}_l\|_F\sqrt{\delta E^k}\right)+ \sqrt{\delta E^L},
\end{equation}
where $\mathbf{\tilde{Y}}^l\!=\!\sigma({\mathbf{\hat{W}}_l}^\top\mathbf{\tilde{Y}}^{l-1})$, for $2\leq l\leq L$ denotes `accumulated pruned output' of layer $l$, and $\mathbf{\tilde{Y}}^1\!=\!\sigma({\mathbf{\hat{W}}_1}^\top\mathbf{X})$.
\end{theorem}
Theorem \ref{th1} shows that: 1) Layer-wise error for a layer $l$ will be scaled by continued multiplication of parameters' Frobenius Norm over the following layers when it propagates to final output, i.e., the $L\!-\!l$ layers after the $l$-th layer; 2) The final error of ultimate network output is bounded by the weighted sum of layer-wise errors. The proof of Theorem~\ref{th1} can be found in Appendix.

Consider a general case with (\ref{changel}) and (\ref{finallayer}): parameter $\boldsymbol{\Theta}_{l_{[q]}}$ who has the smallest sensitivity in layer $l$ is pruned by the $i$-th pruning operation, and this finally adds $\prod_{k=l+1}^L\|\hat{\Theta}_k\|_F\sqrt{\delta E^l}$ to the ultimate network output error. It is worth to mention that although it seems that the layer-wise error is scaled by a quite large product factor, $S_l = \prod_{k=l+1}^L\|\hat{\Theta}_k\|_F$ when it propagates to the final layer, this scaling is still tractable in practice because ultimate network output is also scaled by the same product factor compared with the output of layer $l$. For example, we can easily estimate the norm of ultimate network output via, $\|\mathbf{Y}^{L}\|_F \approx S_1\|\mathbf{Y}^{1}\|_F$. If one pruning operation in the 1st layer causes the layer-wise error $\sqrt{\delta E^1}$, then the \textbf{relative} ultimate output error is
$$\xi_r^L=\frac{\|\mathbf{\tilde{Y}}^L - \mathbf{Y}^{L}\|_F}{\|\mathbf{Y}^{L}\|_F} \approx \frac{\sqrt{\delta E^1}}{\|\frac{1}{n}\mathbf{Y}^{1}\|_F}.$$
Thus, we can see that even $S_1$ may be quite large, the relative ultimate output error would still be about $\sqrt{\delta E^1}/\|\frac{1}{n}\mathbf{Y}^{1}\|_F$ which is controllable in practice especially when most of modern deep networks adopt maxout layer \cite{maxout} as ultimate output. Actually, $S_0$ is called as network gain representing the ratio of the magnitude of the network output to the magnitude of the network input.

\subsection{The Proposed Algorithm}
\label{parallel}
\subsubsection{Pruning on Fully-Connected Layers}
To selectively prune parameters, our approach needs to compute the inverse Hessian matrix at each layer to measure the sensitivities of each parameter of the layer, which is still computationally expensive though tractable. In this section, we present an efficient algorithm that can reduce the size of the Hessian matrix and thus speed up computation on its inverse.

\begin{figure}[t!]
\begin{center}
\centerline{\includegraphics[width=0.5\textwidth]{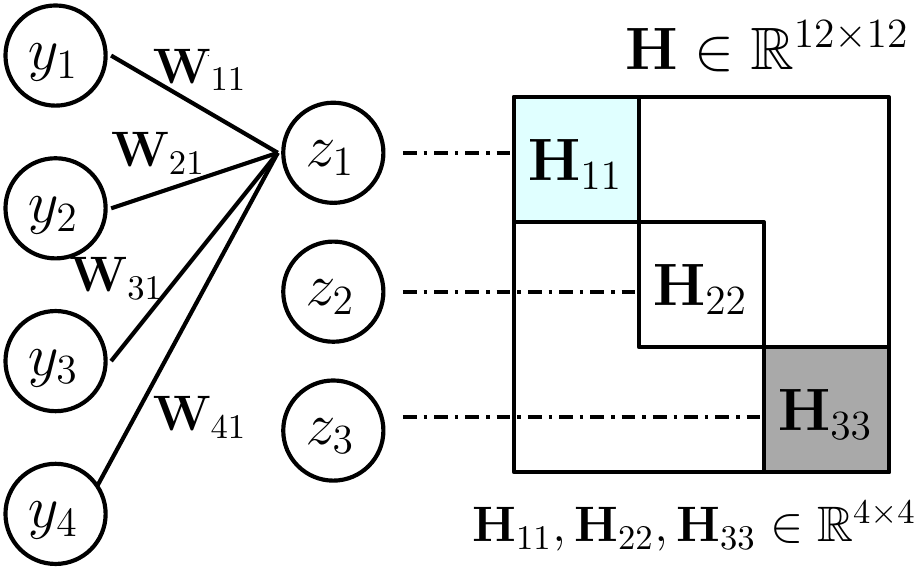}}
\caption{Illustration of shape of Hessian. For feed-forward neural networks, unit $z_1$ gets its activation via forward propagation:
    $\mathbf{z}=\mathbf{W}^\top\mathbf{y}$, where $\mathbf{W}\in \mathbb{R}^{4\times 3}$, $\mathbf{y}=[y_1, y_2, y_3, y_4]^\top\in \mathbb{R}^{4\times 1}$, and $\mathbf{z}=[z_1, z_2, z_3]^\top\in \mathbb{R}^{3\times 1}$.
    Then the Hessian matrix of $z_1$ w.r.t. all parameters is denoted by $\mathbf{H}^{[z_1]}$. As illustrated in the figure, $\mathbf{H}^{[z_1]}$'s elements are zero except for those corresponding to $\mathbf{W}_{*1}$ (the 1st column of $\mathbf{W}$), which is denoted by $\mathbf{H}_{11}$. $\mathbf{H}^{[z_2]}$ and $\mathbf{H}^{[z_3]}$ are similar. More importantly, $\mathbf{H}^{-1}=\mbox{diag}(\mathbf{H}_{11}^{-1},\mathbf{H}_{22}^{-1},\mathbf{H}_{33}^{-1})$, and $\mathbf{H}_{11}=\mathbf{H}_{22}=\mathbf{H}_{33}$. As a result, one only needs to compute $\mathbf{H}_{11}^{-1}$ to obtain $\mathbf{H}^{-1}$ which significantly reduces computational complexity.}
\label{fig:illu}
\end{center}
\end{figure}

For each layer $l$, according to the definition of the error function used in Lemma \ref{th11}, the first derivative of the error function with respect to $\boldsymbol{\hat{\Theta}}_l$ is
$\frac{\partial E^l}{\partial \boldsymbol{\Theta}_l} = -\frac{1}{n}\sum_{j=1}^n\frac{\partial z_j^l}{\partial \boldsymbol{\Theta}_l}(\mathbf{\hat{z}}_j^l - \mathbf{z}_j^l)$,
where $\mathbf{\hat{z}}_j^l$ and $\mathbf{z}_j^l$ are the $j$-th columns of the matrices $\mathbf{\hat{Z}}^l$ and $\mathbf{Z}^l$, respectively, and the Hessian matrix is defined as:
$\mathbf{H}_l \!\equiv\! \frac{\partial^2 E^l}{\partial {\left(\boldsymbol{\Theta}_l\right)}^2} \!=\! \frac{1}{n}\sum_{j=1}^n \left(\frac{\partial z_j^l}{\partial \boldsymbol{\Theta}_l} {\left(\frac{\partial z_j^l}{\partial \boldsymbol{\Theta}_l}\right)}^\top \!-\! \frac{\partial^2 z_j^l}{\partial {(\boldsymbol{\Theta}_l)}^2}(\mathbf{\hat{z}}_j^l \!-\! \mathbf{z}_j^l)^\top \right)$.
Note that for most cases $\mathbf{\hat{z}}_j^l$ is quite close to $\mathbf{z}_j^l$, we simply ignore the term containing $\mathbf{\hat{z}}_j^l \!-\! \mathbf{{z}}_j^l$. Even in the late-stage of pruning when this difference is not small, we can still ignore the corresponding term~\cite{obs}. For layer $l$ that has $m_l$ output units, $\mathbf{z}_j^l \!=\! [z_{1j}^l,\ldots, z_{m_lj}^l]$, the Hessian matrix can be calculated via
\begin{equation}
\label{finalhessian}
	\mathbf{H}_l = \frac{1}{n}\sum_{j=1}^n \mathbf{H}^j_l=\frac{1}{n}\sum_{j=1}^n\sum_{i=1}^{m_l}\frac{\partial z_{ij}^l}{\partial \boldsymbol{\Theta}_l} {\left( \frac{\partial z_{ij}^l}{\partial \boldsymbol{\Theta}_l} \right)}^\top,
\end{equation}
where the Hessian matrix for a single instance $j$ at layer $l$, $\mathbf{H}^j_l$, is a block diagonal square matrix of the size $m_{l-1}\!\times\! m_l$. Specifically, the gradient of the first output unit $z_{1j}^l$ w.s.t. $\boldsymbol{\Theta}_l$ is
$\frac{\partial z_{1j}^l}{\partial \boldsymbol{\Theta}_l} \!=\! \left[ \frac{\partial z_{1j}^l}{\partial \mathbf{w}_{1}}, \ldots, \frac{\partial z_{1j}^l}{\partial \mathbf{w}_{m_l}} \right]$,
where $\mathbf{w}_i$ is the $i$-th column of $\mathbf{W}_l$. As $z_{1j}^l$ is the layer output before activation function, its gradient is simply to calculate, and more importantly all output units's gradients are equal to the layer input: $\frac{\partial z_{ij}^l}{\partial \mathbf{w}_{k}} \!=\! \mathbf{y}_j^{l-1}$ if $k=i$, otherwise $\frac{\partial z_{ij}^l}{\partial \mathbf{w}_{k}} \!=\! 0$. An illustrated example is shown in Figure~\ref{fig:illu}, where we ignore the scripts $j$ and $l$ for simplicity in presentation.

It can be shown that the block diagonal square matrix $\mathbf{H}^j_l$'s diagonal blocks $\mathbf{H}^j_{l_{ii}}\in\mathbb{R}^{m_{l-1}\times m_{l-1}}$, where $1\leq i\leq m_l$, are all equal to $\boldsymbol{\psi}^j_l \!=\! \mathbf{y}_j^{l-1}{(\mathbf{y}_j^{l-1})}^\top$, and the inverse Hessian matrix $\mathbf{H}^{-1}_l$ is also a block diagonal square matrix with its diagonal blocks being $(\frac{1}{n}\sum_{j=1}^n\boldsymbol{\psi}^j_l)^{-1}$. In addition, normally $\boldsymbol{\Psi}^l=\frac{1}{n}\sum_{j=1}^n\boldsymbol{\psi}^j_l$ is degenerate and its pseudo-inverse can be calculated recursively via %standard matrix inversion:
Woodbury matrix identity~\cite{obs}:
$${(\boldsymbol{\Psi}^l_{j+1})}^{-1} = {(\boldsymbol{\Psi}^l_j)}^{-1} - \frac{{(\boldsymbol{\Psi}^l_j)}^{-1} \mathbf{y}_{j}^{l-1} {\left(\mathbf{y}_{j}^{l-1}\right)}^\top {(\boldsymbol{\Psi}^l_j)}^{-1}}
{n+{\left(\mathbf{y}_{j+1}^{l-1}\right)}^\top{(\boldsymbol{\Psi}^l_j)}^{-1} \mathbf{y}_{j+1}^{l-1}},$$
where $\boldsymbol{\Psi}^l_t \!=\! \frac{1}{t}\sum_{j=1}^t\boldsymbol{\psi}^j_l$ with ${(\boldsymbol{\Psi}_0^l)}^{-1} \!=\! \alpha \mathbf{I}$, $\alpha\in[{10}^4,{10}^8]$, and ${(\boldsymbol{\Psi}^l)}^{-1} \!=\! {(\boldsymbol{\Psi}_n^l)}^{-1}$. The size of $\boldsymbol{\Psi}^l$ is then reduced to $m_{l-1}$, and the computational complexity of calculating $\mathbf{H}_l^{-1}$ is $O\left(nm_{l-1}^2\right)$.

To make the estimated minimal change of the error function optimal in (\ref{changel}), the layer-wise Hessian matrices need to be exact. Since the layer-wise Hessian matrices only depend on the corresponding layer inputs, they are always able to be exact even after several pruning operations. The only parameter we need to control is the layer-wise error $\varepsilon^l$. Note that there may be a ``pruning inflection point'' after which layer-wise error would drop dramatically. In practice, user can incrementally increase the size of pruned parameters based on the sensitivity $L_q$, and make a trade-off between the pruning ratio and the performance drop to set a proper tolerable error threshold or pruning ratio.

The procedure of our pruning algorithm for a fully-connected layer $l$ is summarized as follows.
\begin{enumerate}
	\item[Step 1:] Get layer input $\mathbf{y}^{l-1}$ from a well-trained deep network.
	\item[Step 2:] Calculate the Hessian matrix $\mathbf{H}_{l_{ii}}$, for $i=1,...,m_l$, and its pseudo-inverse over the dataset, and get the whole pseudo-inverse of the Hessian matrix.
	\item[Step 3:] Compute optimal parameter change $\delta \boldsymbol{\Theta}_l$ and the sensitivity $L_q$ for each parameter at layer $l$. Set tolerable error threshold $\epsilon$.
	\item[Step 4:] Pick up parameters $\boldsymbol{\Theta}_{l_{[q]}}$'s with the smallest sensitivity scores.
	\item[Step 5:] If $\sqrt{L_{q}}\leq \epsilon$, prune the parameter $\boldsymbol{\Theta}_{l_{[q]}}$'s and get new parameter values via $\boldsymbol{\hat{\Theta}}_l=\boldsymbol{\Theta}_l+\delta \boldsymbol{\Theta}_l$, then repeat Step 4; otherwise stop pruning.
\end{enumerate}

\subsubsection{Pruning on Convolutional Layers}
It is straightforward to generalize our method to a convolutional layer and its variants if we vectorize filters of each channel and consider them as special fully-connected layers that have multiple inputs (patches) from a single instance. Consider a vectorized filter $\mathbf{w}_i$ of channel $i$, $1\leq i\leq m_l$, it acts similarly to parameters which are connected to the same output unit in a fully-connected layer. However, the difference is that for a single input instance $j$, every filter step of a sliding window across of it will extract a patch $C_{j_n}$ from the input volume. Similarly, each pixel $z^l_{ij_n}$ in the 2-dimensional activation map that gives the response to each patch corresponds to one output unit in a fully-connected layer. Hence, for convolutional layers, (\ref{finalhessian}) is generalized as $\mathbf{H}_l=\frac{1}{n}\sum_{j=1}^n\sum_{i=1}^{m_l}\sum_{j_n}\frac{\partial z^l_{ij_n}}{\partial [\mathbf{w}_1, \ldots, \mathbf{w}_{m_l}]}$,
where $\mathbf{H}_l$ is a block diagonal square matrix whose diagonal blocks are all the same. Then, we can slightly revise the computation of the Hessian matrix, and extend the algorithm for fully-connected layers to convolutional layers.

Note that the accumulated error of ultimate network output can be linearly bounded by layer-wise error as long as the model is feed-forward. Thus, L-OBS is a general pruning method and friendly with most of feed-forward neural networks whose layer-wise Hessian can be computed expediently with slight modifications. However, if models have sizable layers like ResNet-101, L-OBS may not be economical because of computational cost of Hessian, which will be studied in our future work.

\section{Experiments}
%\subsection{Experimental Setup}
In this section, we verify the effectiveness of our proposed Layer-wise OBS (L-OBS) using various architectures of deep neural networks in terms of compression ratio (CR), error rate before retraining, and the number of iterations required for retraining to resume satisfactory performance. CR is defined as the ratio of the number of preserved parameters to that of original parameters, lower is better. We conduct comparison results of L-OBS with the following pruning approaches: 1) Randomly pruning, 2) OBD~\cite{obd}, 3) LWC~\cite{hannips}, 4) DNS~\cite{dynamic}, and 5) Net-Trim~\cite{nettrim}. The deep architectures used for experiments include: LeNet-300-100~\cite{lenet5} and LeNet-5~\cite{lenet5} on the MNIST dataset, CIFAR-Net\footnote{A revised AlexNet for CIFAR-10 containing three convolutional layers and two fully connected layers.}~\cite{cifarnet} on the CIFAR-10 dataset, AlexNet~\cite{alexnet} and VGG-16~\cite{vgg16} on the ImageNet ILSVRC-2012 dataset.
%More specifically, LeNet-300-100 is a classical feed-forward network, which has three fully connected layers, with 267K learnable parameters. LeNet-5 is a convolutional neural network that has two convolutional layers and two fully connected layers. CIFAR-Net is a revised AlexNet for CIFAR-10 containing three convolutional layers and two fully connected layers. AlexNet contains 61M parameters across 5 convolutional layers and 3 fully-connected layers and a larger. VGG-16 contains 138M parameters across 13 convolutional layers and 3 fully-connected layers.
For experiments, we first well-train the networks, and apply various pruning approaches on networks to evaluate their performance. The retraining batch size, crop method and other hyper-parameters are under the same setting as used in LWC. Note that to make comparisons fair, we do not adopt any other pruning related methods like Dropout or sparse regularizers on MNIST. In practice, L-OBS can work well along with these techniques as shown on CIFAR-10 and ImageNet.

\subsection{Overall Comparison Results}
\label{sec:exp:overall}
The overall comparison results are shown in Table~\ref{sample-table}. In the first set of experiments, we prune each layer of the well-trained LeNet-300-100 with compression ratios: 6.7\%, 20\% and 65\%, achieving slightly better overall compression ratio (7\%) than LWC (8\%). Under comparable compression ratio, L-OBS has quite less drop of performance (before retraining) and lighter retraining compared with LWC whose performance is almost ruined by pruning. Classic pruning approach OBD is also compared though we observe that Hessian matrices of most modern deep models are strongly non-diagonal in practice. Besides relative heavy cost to obtain the second derivatives via the chain rule, OBD suffers from drastic drop of performance when it is directly applied to modern deep models.

To properly prune each layer of LeNet-5, we increase tolerable error threshold $\epsilon$ from relative small initial value to incrementally prune more parameters, monitor model performance, stop pruning and set $\epsilon$ until encounter the ``pruning inflection point'' mentioned in Section \ref{parallel}. In practice, we prune each layer of LeNet-5 with compression ratio: 54\%, 43\%, 6\% and 25\% and retrain pruned model with much fewer iterations compared with other methods (around $1\!:\!1000$). As DNS retrains the pruned network after every pruning operation, we are not able to report its error rate of the pruned network before retraining. However, as can be seen, similar to LWC, the total number of iterations used by DNS for rebooting the network is very large compared with L-OBS. Results of retraining iterations of DNS are reported from~\cite{dynamic} and the other experiments are implemented based on TensorFlow~\cite{tensorflow}. In addition, in the scenario of requiring high pruning ratio, L-OBS can be quite flexibly adopted to an iterative version, which performs pruning and light retraining alternatively to obtain higher pruning ratio with relative higher cost of pruning. With two iterations of pruning and retraining, L-OBS is able to achieve as the same pruning ratio as DNS with much lighter total retraining: 643 iterations on LeNet-300-100 and 841 iterations on LeNet-5.

Regarding comparison experiments on CIFAR-Net, we first well-train it to achieve a testing error of 18.57\% with Dropout and Batch-Normalization. We then prune the well-trained network with LWC and L-OBS, and get the similar results as those on other network architectures. We also observe that LWC and other retraining-required methods always require much smaller learning rate in retraining. This is because representation capability of the pruned networks which have much fewer parameters is damaged during pruning based on a principle that number of parameters is an important factor for representation capability. However, L-OBS can still adopt original learning rate to retrain the pruned networks. Under this consideration, L-OBS not only ensures a warm-start for retraining, but also finds important connections (parameters) and preserve capability of representation for the pruned network instead of ruining model with pruning.

\begin{table}[t!]
\centering
\scriptsize
\ExerciseCaption{Overall comparison results. (For iterative L-OBS, err. after pruning regards the last pruning stage.)}
\label{sample-table}
\begin{tabular}{lccccccr}
\toprule
\toprule
Method & Networks & Original error & CR & \!\!Err. after pruning\!\! & Re-Error & \#Re-Iters. 	\\
\midrule
Random      & LeNet-300-100 &  1.76\% & 8\% & 85.72\%  & 2.25\% & $3.50\times 10^5$ \\
OBD         & LeNet-300-100 &  1.76\% & 8\% & 86.72\%  & 1.96\% & $8.10\times 10^4$ \\
LWC         & LeNet-300-100 &  1.76\% & 8\% & 81.32\%  & 1.95\% & $1.40\times 10^5$\\
DNS         & LeNet-300-100 &  1.76\% & 1.8\% & -      & 1.99\% & $3.40\times 10^4$\\
L-OBS       & LeNet-300-100 & 1.76\% & 7\% &\textbf{3.10\%}& \text{1.82\%} & \textbf{510} \\
L-OBS (iterative)\!\!  & LeNet-300-100 & 1.76\% & \textbf{1.5\%} & 2.43\% & \text{1.96\%} & \textbf{643} \\
\midrule
OBD         & LeNet-5  & 1.27\%   & 8\% & 86.72\%       & 2.65\% & $2.90\times 10^5$\\
LWC         & LeNet-5  & 1.27\%   & 8\% & 89.55\%       & 1.36\% & $9.60\times 10^4$\\
DNS         & LeNet-5  & 1.27\%   & \textbf{0.9\%}      & -       & 1.36\% & $4.70\times 10^4$\\
L-OBS       & LeNet-5  & 1.27\%   & 7\% &\textbf{3.21\%}& 1.27\% & \textbf{740} \\
L-OBS (iterative)\!\! & LeNet-5  & 1.27\%   & \textbf{0.9\%} &2.04\%& 1.66\% & \textbf{841} \\
\midrule
LWC         & CIFAR-Net &  18.57\% & 9\% & 87.65\%  & 19.36\%    &   $1.62\times 10^5$    \\
L-OBS       & CIFAR-Net &  18.57\% & 9\% & \textbf{21.32\%}  & 18.76\%    &   \textbf{1020}    \\
\midrule
DNS         & \!\!AlexNet (Top-1 / Top-5 err.)\!\!      & 43.30 / 20.08\% &\textbf{5.7\%} & -       & 43.91 / 20.72\% & $7.30\times 10^5$\\
LWC         & AlexNet (Top-1 / Top-5 err.)      & 43.30 / 20.08\% &11\%           & 76.14 / 57.68\% & 44.06 / 20.64\% & $5.04\times 10^6$\\
L-OBS       & AlexNet (Top-1 / Top-5 err.)      & 43.30 / 20.08\% &11\%           & \textbf{50.04 / 26.87\%}& 43.11 / 20.01\% & \!\!$\mathbf{1.81\times 10^4}$\!\! \\
\midrule
DNS & VGG-16 (Top-1 / Top-5 err.)     & 31.66 / 10.12\% &7.5\%           & - & \!\!63.38\% / 38.69\%\!\! & ~$1.07\times 10^6$\\
LWC         & VGG-16 (Top-1 / Top-5 err.)     & 31.66 / 10.12\% &7.5\%           & 73.61 / 52.64\% & 32.43 / 11.12\% & $2.35\times 10^7$\\
L-OBS (iterative)       & VGG-16 (Top-1 / Top-5 err.)      & 31.66 / 10.12\% &7.5\%           & \textbf{37.32 / 14.82\%}& 32.02 / 10.97\% & \!\!$\mathbf{8.63\times 10^4}$\!\! \\
\bottomrule
\end{tabular}
\end{table}

Regarding AlexNet, L-OBS achieves an overall compression ratio of 11\% without loss of accuracy with 2.9 hours on 48 Intel Xeon(R) CPU E5-1650 to compute Hessians and 3.1 hours on NVIDIA Tian X GPU to retrain pruned model (i.e. 18.1K iterations). The computation cost of the Hessian inverse in L-OBS is negligible compared with that on heavy retraining in other methods. This claim can also be supported by the analysis of time complexity. As mentioned in Section~\ref{parallel}, the time complexity of calculating $\mathbf{H}_l^{-1}$ is $O\left(nm_{l-1}^2\right)$. Assume that neural networks are retrained via SGD, then the approximate time complexity of retraining is $O\left(IdM\right)$, where $d$ is the size of the mini-batch, $M$ and $I$ are the total numbers of parameters and iterations, respectively. By considering that $M\approx\sum_{l=1}^{l=L}\left(m^2_{l-1}\right)$, and retraining in other methods always requires millions of iterations ($Id\gg n$) as shown in experiments, complexity of calculating the Hessian (inverse) in L-OBS is quite economic. More interestingly, there is a trade-off between compression ratio and pruning (including retraining) cost. Compared with other methods, L-OBS is able to provide fast-compression: prune AlexNet to 16\% of its original size without substantively impacting accuracy (pruned top-5 error 20.98\%) even without any retraining. We further apply L-OBS to VGG-16 that has 138M parameters. To achieve more promising compression ratio, we perform pruning and retraining alteratively twice. As can be seen from the table, L-OBS achieves an overall compression ratio of 7.5\% without loss of accuracy taking 10.2 hours in total on 48 Intel Xeon(R) CPU E5-1650 to compute the Hessian inverses and 86.3K iterations to retrain the pruned model.

\begin{figure}[t!]
	\center
	\subfigure[Top-5 test accuracy of L-OBS on ResNet-50 under different compression ratios.]
	{\label{fig:resnet}\includegraphics[width=0.45\columnwidth]{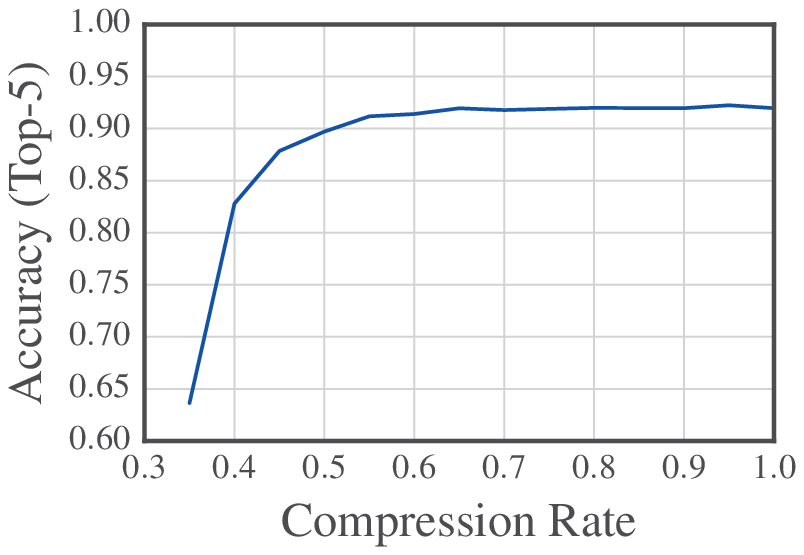}}\hspace{5mm}
	\subfigure[Memory Comparion between L-OBS and Net-Trim on MNIST.]
	{\label{fig:net-trim}\includegraphics[width=0.45\textwidth]{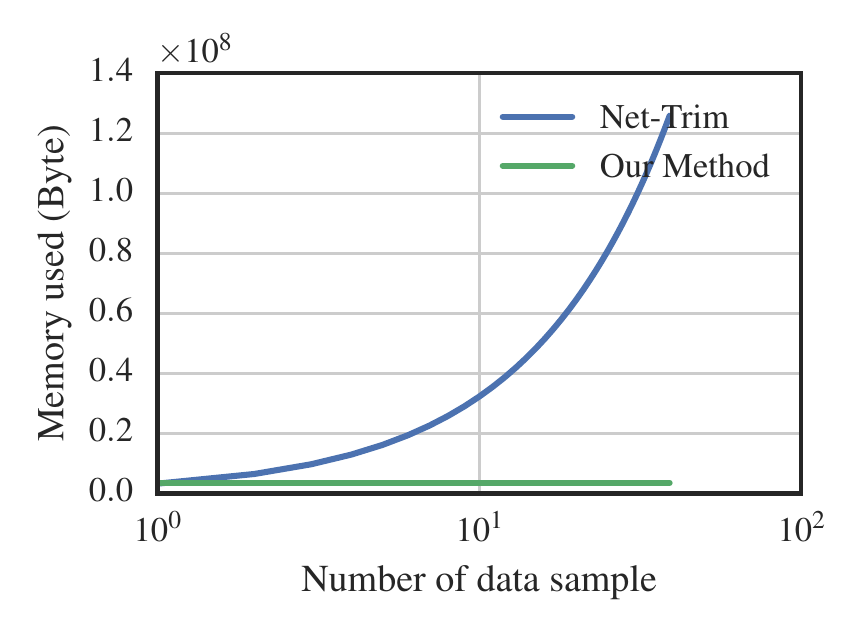}}\vspace{-3mm}
\end{figure}

%\subsection{Experiment Result on ResNet-50}
We also apply L-OBS on ResNet-50~\cite{he2016deep}. From our best knowledge, this is the first work to perform pruning on ResNet. %ResNet-50 contains 53 convolutional layers and 1 fully-connected layer.
We perform pruning on all the layers: All layers share a same compression ratio, and we change this compression ratio in each experiments. The results are shown in Figure~\ref{fig:resnet}. As we can see, L-OBS is able to maintain ResNet's accuracy (above 85\%) when the compression ratio is larger than or equal to 45\%.

\begin{table}[t!]
\ExerciseCaption{Comparison of Net-Trim and Layer-wise OBS on the second layer of LeNet-300-100.}
\label{nettrimtable}
\begin{center}
\scriptsize
\begin{tabular}{lccrlccr}
\toprule
\toprule
Method & $\xi^2_r$ & Pruned Error & CR &Method & $\xi^2_r$ & Pruned Error & CR\\
\midrule
Net-Trim& 0.13  &  13.24\%  &  19\%  & Net-Trim & 0.62  &  28.45\%  &  7.4\%\\

L-OBS   & 0.70  &  11.34\%   &   3.4\% & L-OBS   & 0.37  &  4.56\%   &   7.4\%   \\
L-OBS   & 0.71  &  10.83\%   &   3.8\% & Net-Trim   & 0.71  &  47.69\%   &   4.2\%    \\
\bottomrule
\end{tabular}\vspace{-3mm}
\end{center}
\end{table}

\vspace{-1mm}
\subsection{Comparison between L-OBS and Net-Trim}
As our proposed L-OBS is inspired by Net-Trim, which adopts $\ell_1$-norm to induce sparsity, we conduct comparison experiments between these two methods. In Net-Trim, networks are pruned by formulating layer-wise pruning as a optimization: $\min_{\mathbf{W}_l}\|\mathbf{W}_l\|_1$ s.t. $\|\sigma(\mathbf{W}_l^\top \mathbf{Y}^{l-1})-\mathbf{Y}^l\|_F\leq \xi^l$, where $\xi^l$ corresponds to $\xi^l_r\|\mathbf{Y}^l\|_F$ in L-OBS. Due to memory limitation of Net-Trim, we only prune the middle layer of LeNet-300-100 with L-OBS and Net-Trim under the same setting. As shown in Table~\ref{nettrimtable}, under the same pruned error rate, CR of L-OBS outnumbers that of the Net-Trim by about six times. In addition, Net-Trim encounters explosion of memory and time on large-scale datasets and large-size parameters. Specifically, space complexity of the positive semidefinite matrix $Q$ in quadratic constraints used in Net-Trim for optimization is $O\left(2nm_l^2m_{l-1}\right)$. For example, $Q$ requires about 65.7Gb for 1,000 samples on MNIST as illustrated in Figure~\ref{fig:net-trim}. Moreover, Net-Trim is designed for multi-layer perceptrons and not clear how to deploy it on convolutional layers.

\vspace{-3mm}
\section{Conclusion}
We have proposed a novel L-OBS pruning framework to prune parameters based on second order derivatives information of the layer-wise error function and provided a theoretical guarantee on the overall error in terms of the reconstructed errors for each layer. Our proposed L-OBS can prune considerable number of parameters with tiny drop of performance and reduce or even omit retraining. More importantly, it identifies and preserves the real important part of networks when pruning compared with previous methods, which may help to dive into nature of neural networks.
\vspace{-3mm}
\section*{Acknowledgements}
This work is supported by NTU Singapore Nanyang Assistant Professorship (NAP) grant M4081532.020, Singapore MOE AcRF Tier-2 grant MOE2016-T2-2-060, and Singapore MOE AcRF Tier-1 grant 2016-T1-001-159.


\bibliographystyle{unsrt}
%\bibliography{example_paper}

\newpage
%%%%%%%%%%%%%%%%%%%%%%%%%%%%%%%%%%%5
%%%%%%%%%%%%%%%%%%%%%%%%%%%%%%%%%%%%

\section*{APPENDIX}
\label{introduction}
%Inspired by the structure of human brain, deep neural networks have become a ubiquitous technology in many real-world applications~\cite{deepnature}.
\section*{Proof of Theorem~\ref{th1}}

We prove Theorem~\ref{th1} via induction. First, for $l\!=\!1$, (\ref{finallayer}) holds as a special case of (\ref{onelayer}). Then suppose that Theorem \ref{th1} holds up to layer $l$:
\begin{equation}
\label{firstone}
	\tilde{\varepsilon}^l\leq \sum_{h=1}^{l-1}(\prod_{k=h+1}^l\|\boldsymbol{\hat{\Theta}}_k\|_F\sqrt{\delta E^h})+ \sqrt{\delta E^l}
\end{equation}
In order to show that (\ref{firstone}) holds for layer $l+1$ as well, we refer to $\mathbf{\hat{Y}}^{l+1}\!=\!\sigma(\mathbf{\hat{W}}_{l+1}^\top \mathbf{Y}^{l})$ as `layer-wise pruned output', where the input $\mathbf{Y}^{l}$ is fixed as the same as the originally well-trained network not an accumulated input $\mathbf{\tilde{Y}}^{l}$, and have the following theorem.

\begin{theorem}
Consider layer $l\!+\!1$ in a pruned deep network, the difference between its accumulated pruned output, $\mathbf{\tilde{Y}}^{l+1}$, and layer-wise pruned output, $\mathbf{\hat{Y}}^{l+1}$, is bounded by:
\begin{equation}
\label{wiseandaccu}
	\|\mathbf{\tilde{Y}}^{l+1}-\mathbf{\hat{Y}}^{l+1}\|_F\leq %\sqrt{n}\|\textcolor{red}{\mathbf{\hat{\Theta}}^{l+1}}\|_F \tilde{\varepsilon}^l.
	\sqrt{n}\|\mathbf{\hat{\Theta}}^{l+1}\|_F \tilde{\varepsilon}^l.
\end{equation}
\end{theorem}
\textit{Proof sketch:} Consider one arbitrary element of the layer-wise pruned output $\mathbf{\hat{Y}}^{l+1}$:
\begin{eqnarray*}
	\hat{y}^{l+1}_{ij} &=& \sigma(\mathbf{\hat{w}}_i^\top\mathbf{\tilde{y}}_j^{l} + \mathbf{\hat{w}}_i^\top(\mathbf{y}_j^{l}-\mathbf{\tilde{y}}_j^{l}))\\
                        &\leq & \mathbf{\tilde{y}}^{l+1}_{ij} + \sigma(\mathbf{\hat{w}}_i^\top(\mathbf{y}_j^{l} - \mathbf{\tilde{y}}_j^{l}))\\
	                    &\leq & \mathbf{\tilde{y}}^{l+1}_{ij} + \vert\mathbf{\hat{w}}_i^\top(\mathbf{y}_j^{l} - \mathbf{\tilde{y}}_j^{l})\vert,
\end{eqnarray*}
where $\mathbf{\hat{w}}_i$ is the $i$-th column of $\mathbf{\hat{W}}_{l+1}$. The first inequality is obtained because we suppose the activation function $\sigma(\cdot)$ is ReLU. Similarly, it holds for accumulated pruned output:
\[\tilde{y}^{l+1}_{ij}\leq \hat{y}^{l+1}_{ij} + \vert\mathbf{\hat{w}}_i^\top(\mathbf{y}_j^{l}-\mathbf{\tilde{y}}_j^{l})\vert.\]
By combining the above two inequalities, we have $$\vert\tilde{y}^{l+1}_{ij}-\hat{y}^{l+1}_{ij}\vert \leq \vert\mathbf{\hat{w}}_i^\top(\mathbf{y}_j^{l}-\mathbf{\tilde{y}}_j^{l})\vert,$$ and thus have the following inequality in a form of matrix,
\begin{eqnarray*}
\|\mathbf{\tilde{Y}}^{l+1} - \mathbf{\hat{Y}}^{l+1}\|_F
    \leq  \|\mathbf{\hat{W}}_{l+1} (\mathbf{Y}^{l} - \mathbf{\tilde{Y}}^{l})\|_F
	\leq  \|\mathbf{\hat{\Theta}}^{l+1}\|_F\|\mathbf{Y}^{l}-\mathbf{\tilde{Y}}^{l}\|_F
\end{eqnarray*}
As $\tilde{\varepsilon}^l$ is defined as $\tilde{\varepsilon}^l=\frac{1}{\sqrt{n}}\|\mathbf{Y}^{l}-\mathbf{\tilde{Y}}^{l}\|_F$, we have
\[\|\mathbf{\tilde{Y}}^{l+1} - \mathbf{\hat{Y}}^{l+1}\|_F \leq \sqrt{n}\|\mathbf{\hat{\Theta}}^{l+1}\|_F\tilde{\varepsilon}^l.\]
This completes the proof of Theorem~\ref{wiseandaccu}.

By using (\ref{onelayer}) ,(\ref{wiseandaccu}) and the triangle inequality, we are now able to extend (\ref{firstone}) to layer $l+1$:
%\begin{tiny}
\begin{eqnarray*}
\tilde{\varepsilon}^{l+1}
    = \frac{1}{\sqrt{n}}\|\mathbf{\tilde{Y}}^{l+1} - \mathbf{Y}^{(l+1)}\|_F
	& \leq & \frac{1}{\sqrt{n}}\|\mathbf{\tilde{Y}}^{l+1}-\mathbf{\hat{Y}}^{(l+1)}\|_F + \frac{1}{\sqrt{n}}\|\mathbf{\hat{Y}}^{l+1}-\mathbf{Y}^{(l+1)}\|_F \\
    & \leq & \sum_{h=1}^{l}\left(\prod_{k=h+1}^{l+1}\|\mathbf{\hat{\Theta}}^{k+1}\|_F\cdot\sqrt{\delta E^h}\right)+ \sqrt{\delta E^{l+1}}.
\end{eqnarray*}
%\end{tiny}
Finally, we prove that (\ref{firstone}) holds up for all layers, and Theorem~\ref{th1} is a special case when $l\!=\!L$.

\section*{Extensive Experiments and Details}
\subsection*{Redundancy of Networks}
\label{exp1}
LeNet-300-100 is a classical feed-forward network, which has three fully connected layers, with 267K learnable parameters. LeNet-5 is a convolutional neural network that has two convolutional layers and two fully connected layers, with 431K learnable parameters. CIFAR-Net is a revised AlexNet for CIFAR-10 containing three convolutional layers and two fully connected layers.

We first validate the redundancy of networks and the ability of our proposed Layer-wise OBS to find parameters with the smallest sensitivity scores with LeNet-300-100 on MINIST. In all cases, we first get a well-trained network without dropout or regularization terms. Then, we use four kinds of pruning criteria: Random, LWC~\cite{hannips}, ApoZW, and Layer-wise OBS to prune parameters, and evaluate performance of the whole network after performing every 100 pruning operations. Here, LWC is a magnitude-based criterion proposed in \cite{hannips}, which prunes parameters based on smallest absolute values. ApoZW is a revised version of ApoZ~\cite{cuhk}, which measures the importance of each parameter ${\mathbf{W}_l}_{ij}$ in layer $l$ via $\tau^l_{ij}=\vert\frac{1}{n}\sum_{p=1}^n(\mathbf{y}^{l-1}_{ip} \times {\mathbf{W}_l}_{ij})\vert$. In this way, both magnitude of the parameter and its inputs are taken into consideration.
\begin{figure}[t!]
\begin{center}
\centerline{\includegraphics[width=0.575\columnwidth]{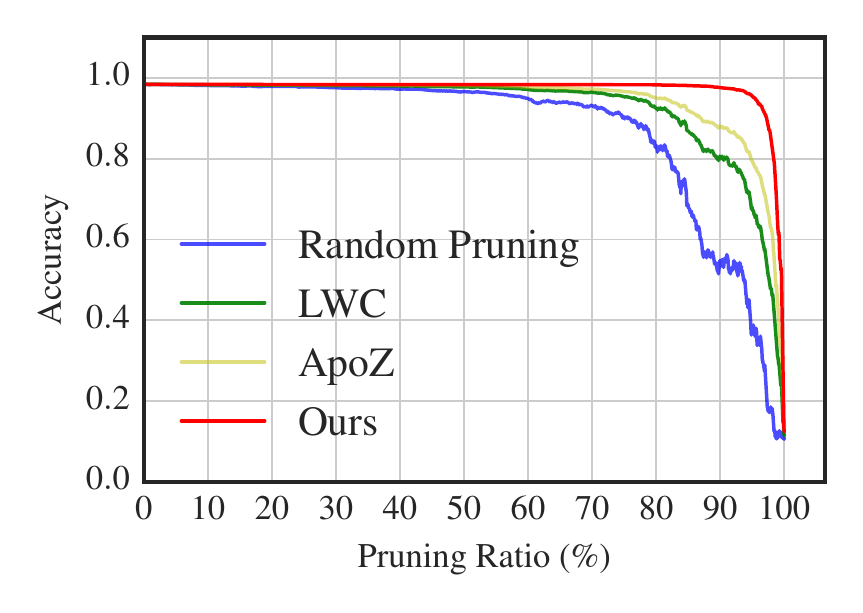}}
\caption{Test accuracy on MNIST using LeNet-300-100 when continually pruning the first layer until pruning ratio is 100\%. Comparison on ability to preserve prediction between LWC, ApoZ and our proposed L-OBS.}
\label{onelayer}
\end{center}\vspace{-4mm}
\end{figure}
\begin{figure}[t!]
\begin{center}
\centerline{\includegraphics[width=0.575\columnwidth]{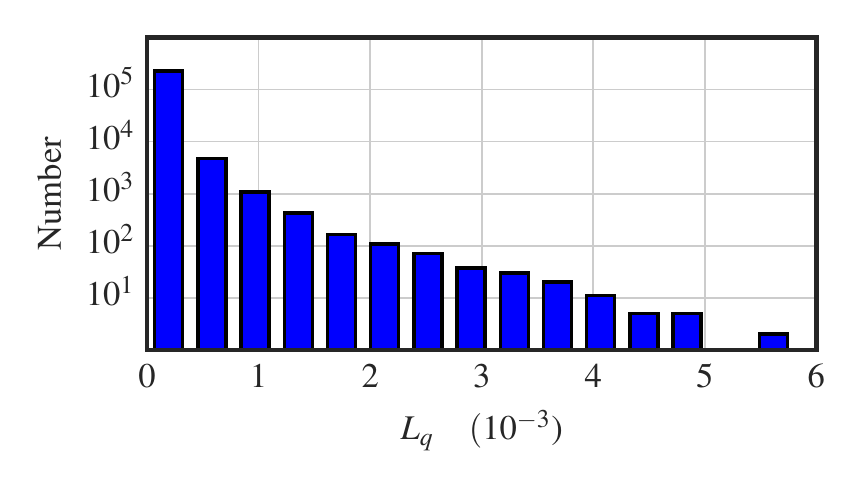}}
\caption{Distribution of sensitivity of parameters in LeNet-300-100's first layer. More than 90\% of parameters' sensitivity scores are smaller than 0.001.}
\label{fig:sensitivity}
\end{center}\vspace{-4mm}
\end{figure}

Originally well-trained model LeNet-300-100 achieves 1.8\% error rate on MNIST without dropout. Four pruning criteria are respectively conducted on the well-trained model's first layer which has 235K parameters by fixing the other two layers' parameters, and test accuracy of the whole network is recorded every 100 pruning operations without any retraining. Overall comparison results are summarized in Figure~\ref{onelayer}.

We also visualize the distribution of parameters' sensitivity scores $L_q$'s estimated by Layer-wise OBS in Figure~\ref{fig:sensitivity}, and find that parameters of little impact on the layer output dominate. This further verifies our hypothesis that deep neural networks usually contain a lot of redundant parameters. As shown in the figure, the distribution of parameters' sensitivity scores in Layer-wise OBS are heavy-tailed. This means that a lot of parameters can be pruned with minor impact on the prediction outcome. Random pruning gets the poorest result as expected but can still preserve prediction accuracy when the pruning ratio is smaller than 30\%. This also indicates the high redundancy of the network.

Compared with LWC and ApoZW, L-OBS is able to preserve original accuracy until pruning ratio reaches about 96\% which we call as ``pruning inflection point''. As mentioned in Section 3.4, the reason on this ``pruning inflection point'' is that the distribution of parameters' sensitivity scores is heavy-tailed and sensitivity scores after ``pruning inflection point'' would be considerable all at once. The percentage of parameters with sensitivity smaller than 0.001 is about 92\% which matches well with pruning ratio at inflection point.

\begin{figure}[t!]
\begin{center}
\centerline{\includegraphics[width=0.575\columnwidth]{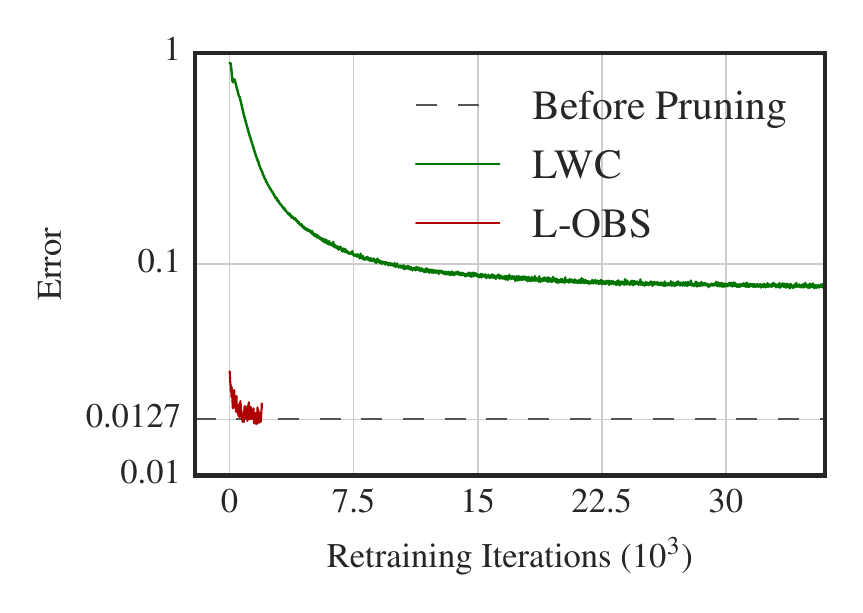}}
\caption{Retraining pattern of LWC and L-OBS. L-OBS has a better start point and totally resume original performance after 740 iterations for LeNet-5.}
\label{err:iter}
\end{center}\vspace{-4mm}
\end{figure}

L-OBS can not only preserve models' performance when pruning one single layer, but also ensures tiny drop of performance when pruning all layers in a model. This claim holds because of the theoretical guarantee on the overall prediction performance of the pruned deep neural network in terms of reconstructed errors for each layer in Section 3.3. As shown in Figure~\ref{err:iter}, L-OBS is able to resume original performance after 740 iterations for LeNet-5 with compression ratio of 7\%.

\subsection*{How To Set Tolerable Error Threshold}
\label{exp2}
One of the most important bounds we proved is that there is a theoretical guarantee on the overall prediction performance of the pruned deep neural network in terms of reconstructed errors for each pruning operation in each layer. This bound enables us to prune a whole model layer by layer without concerns because the accumulated error of ultimate network output is bounded by the weighted sum of layer-wise errors. As long as we control layer-wise errors, we can control the accumulated error.

\begin{figure}[t!]
\begin{center}
\centerline{\includegraphics[width=0.575\columnwidth]{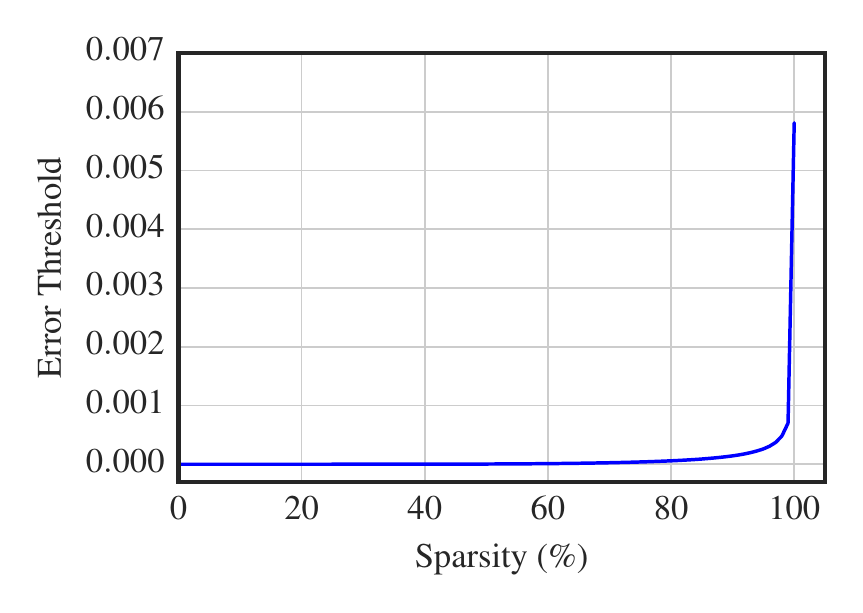}}
\caption{The corresponding relation between tolerable error threshold and pruning ratio.}
\label{thre}
\end{center}\vspace{-4mm}
\end{figure}

Although L-OBS allows users to control the accumulated error of ultimate network output $\tilde{\varepsilon}^L = \frac{1}{\sqrt{n}}\|\mathbf{\tilde{Y}}^{l}-\mathbf{Y}^{l}\|_F$, this error is used to measure difference between network outputs before and after pruning, and is not strictly inversely proportional to the final accuracy. In practice, one can increase tolerable error threshold $\epsilon$ from a relative small initial value to incrementally prune more and more parameters to monitor model performance, and make a trade-off between compression ratio and performance drop. The corresponding relation (in the first layer of LeNet-300-100) between the tolerable error threshold and the pruning ratio is shown in Figure~\ref{thre}.

\subsection*{Iterative Layer-wise OBS}
\label{exp3}
As mentioned in Section 4.1, to achieve better compression ratio, L-OBS can be quite flexibly adopted to its iterative version, which performs pruning and light retraining alternatively. Specifically, the two-stage iterative L-OBS applied to LeNet-300-100, LeNet-5 and VGG-16 in this work follows the following work flow: pre-train a well-trained model $\rightarrow$ prune model $\rightarrow$ retrain the model and reboot performance in a degree $\rightarrow$ prune again $\rightarrow$ lightly retrain model. In practice, if required compression ratio is beyond the ``pruning inflection point'', users have to deploy iterative L-OBS though ultimate compression ratio is not of too much importance. Experimental results are shown in Tabel~\ref{tab:lenet300100},~\ref{tab:lenet5} and~\ref{tab:vgg}, where CR(n) means ratio of the number of preserved parameters to the number of original parameters after the $n$-th pruning.

\begin{table}[h!]
  \caption{For LeNet-300-100, iterative L-OBS(two-stage) achieves compression ratio of 1.5\%}
  \label{tab:lenet300100}
  \centering
  \small
  \begin{tabular}{lccr}
    \toprule
    Layer     & Weights     & CR1 & CR2\\
    \midrule
    fc1&235K&7\%&1\%\\
    fc2&30K&20\%&4\%\\
    fc3&1K&70\%&54\%\\
    \midrule
    Total&266K&8.7\%&1.5\%\\
    \bottomrule
  \end{tabular}
\end{table}

\begin{table}[h!]
  \caption{For LeNet-5, iterative L-OBS(two-stage) achieves compression ratio of 0.9\%}
  \label{tab:lenet5}
  \centering
  \small
  \begin{tabular}{lccr}
    \toprule
    Layer     & Weights     & CR1 & CR2\\
    \midrule
    conv1&0.5K&60\%&20\%\\
    conv2&25K&60\%&1\%\\
    fc1&400K&6\%&0.9\%\\
    fc2&5K&30\%&8\%\\
    \midrule
    Total&431K&9.5\%&0.9\%\\
    \bottomrule
  \end{tabular}
\end{table}

\begin{table}[h!]
  \caption{For VGG-16, iterative L-OBS(two-stage) achieves compression ratio of 7.5\%}
  \label{tab:vgg}
  \centering
  \small
  \begin{tabular}{lccccccccccccr}
    \toprule
    Layer     &   conv1\_1   & conv1\_2 & conv2\_1 & conv2\_2 & conv3\_1 &   conv3\_2   & conv3\_3 & conv4\_1      \\
    \midrule
    Weights &   2K   & 37K  & 74K  & 148K  & 295K  &   590K    & 590K  & 1M     \\
    \midrule
    CR1&   70\%   & 50\%  & 70\%  & 70\%  & 60\%  &   60\%    & 60\%  & 50\% \\
    \midrule
    CR2&   58\%   & 36\%  & 42\%  & 32\%  & 53\%  &   34\%    & 39\%  & 43\%\\
    \midrule
    \toprule
    Layer & conv4\_2 & conv4\_3  & conv5\_1 & conv5\_2 & conv5\_3 & fc6 & fc7 & fc8\\
    \midrule
    Weights&   2M   & 2M & 2M  &  2M & 2M  &   103M    & 17M  & 4M\\
    \midrule
    CR1&  50\%   & 50\%  & 70\%  & 70\%  & 60\%  &   8\%    & 10\%  & 30\%\\
    \midrule
    CR2&  24\%   & 30\%  & 35\%  & 43\%  & 32\%  &   2\%    & 5\%  & 17\%\\
    \bottomrule
  \end{tabular}
\end{table}

\end{document}